\documentclass{article}

\usepackage{microtype}
\usepackage{graphicx}
\usepackage{booktabs} 

\usepackage{amssymb, amsmath, amsthm}

\usepackage{url}
\usepackage[caption=false]{subfig}
\usepackage{mdwlist}
\usepackage{multirow}
\usepackage{amsthm}
\usepackage{color}

\newcommand{\R}{\mathbb{R}}

\newtheorem{theorem}{Theorem}[section]
\newtheorem{proposition}[theorem]{Proposition}

\newtheorem{lemma}[theorem]{Lemma}

\newtheorem{conjecture*}{Conjecture}

\theoremstyle{plain}

\usepackage{hyperref}

\usepackage[accepted]{icml2018}

\icmltitlerunning{DCFNet}

\begin{document}

\twocolumn[
\icmltitle{
DCFNet: Deep Neural Network with Decomposed Convolutional Filters
}

\begin{icmlauthorlist}
\icmlauthor{Qiang Qiu}{duke}
\icmlauthor{Xiuyuan Cheng}{duke}
\icmlauthor{Robert Calderbank}{duke}
\icmlauthor{Guillermo Sapiro}{duke}
\end{icmlauthorlist}

\icmlaffiliation{duke}{Duke University, Durham, North Carolina, USA. Work partially supported by NSF, DoD, NIH and AFOSR}

\icmlcorrespondingauthor{Xiuyuan Cheng}{xiuyuan.cheng@duke.edu}

\icmlkeywords{Convolutional neural network, regularization of deep network, model compression}

\vskip 0.3in
]



\printAffiliationsAndNotice{}  

\begin{abstract}
Filters in a Convolutional Neural Network (CNN) contain model parameters learned from enormous amounts of data.
In this paper,  we suggest to decompose convolutional filters in CNN as a truncated expansion with pre-fixed bases, 
namely the Decomposed Convolutional Filters network (DCFNet), where the expansion coefficients remain learned from data.
Such a structure not only reduces the number of trainable parameters and computation, but also imposes filter regularity by bases truncation. Through extensive experiments, we consistently observe that DCFNet maintains accuracy for image classification tasks
 with a significant reduction of model parameters, particularly with Fourier-Bessel (FB) bases, and even with random bases.
Theoretically, we analyze the representation stability of DCFNet with respect to input variations, and prove representation stability under generic assumptions on the expansion coefficients. The analysis is consistent with the empirical observations.
\end{abstract}

%
%
\section{Introduction}\label{sec:1}

Convolutional Neural Network (CNN) has become one of the most successful computational models 
in machine learning and artificial intelligence.
Remarkable progress has been achieved in the design of successful CNN {\it network structures}, 
 such as the VGG-Net \cite{simonyan2014very}, ResNet \cite{he2016deep}, and DenseNet \cite{huang2016densely}. 
Less attention has been paid to the design of {\it filter structures} in CNNs. 
Filters, namely the weights in the convolutional layers, 
are one of the most important ingredients of a CNN model, 
as filters contain the actual model parameters learned from enormous amounts of data.
Filters in CNNs are typically randomly initialized, 
and then updated using  variants and extensions of gradient descent 
(``back-propagation"). 
As a result, 
trained CNN filters have no specific structures, 
which often leads to significant redundancy in the learned model \cite{DentonZBLF14,han2015learning,SqueezeNet}. 
Filters with improved properties will have a direct impact on the accuracy and efficiency of CNN,
 and the theoretical analysis of filters is also of central importance to the mathematical understanding of deep networks.

\begin{figure}[t]
\vskip 0.2in
\begin{center}
\includegraphics[ width = \linewidth, height=0.35\linewidth]{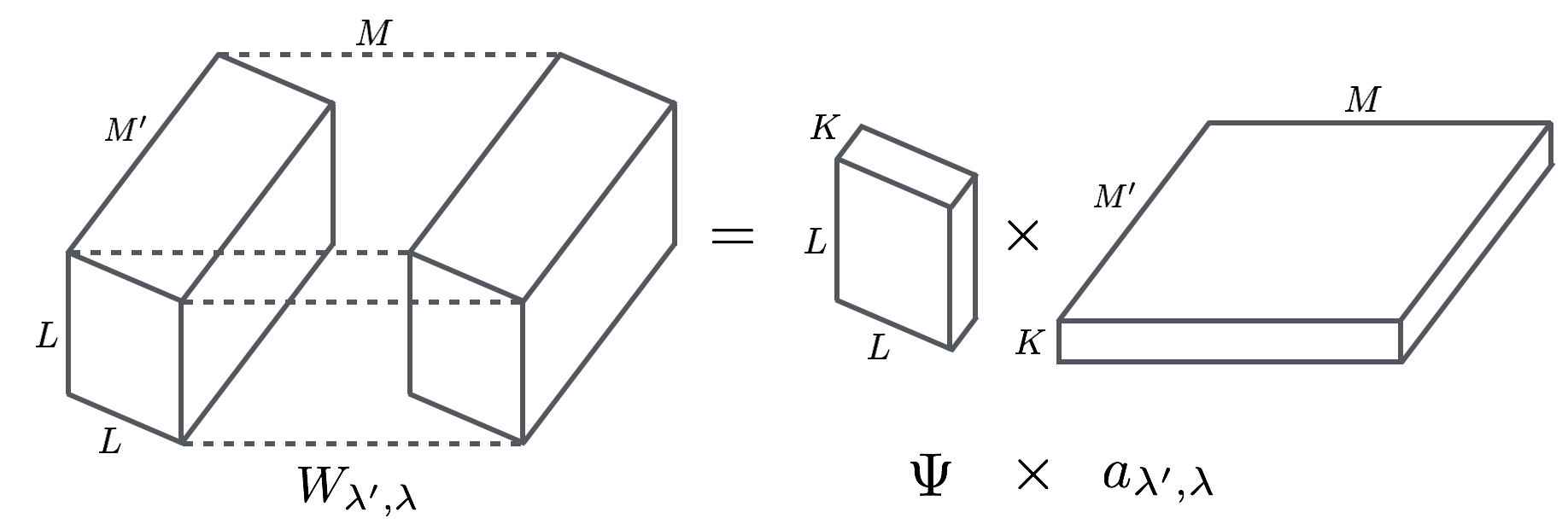} 
 \vskip -0.05in
\caption{
 In a DCFNet, 
 an  $L \times L \times M' \times M$ convolutional layer is decomposed into the
 product of 
 $K$ bases of size $L \times L$ ($\Psi$)
 and $K M' \times M$ coefficients ($a$),
 where $\Psi$ is pre-fixed, and $a$ is learned from data.
 The basis can carry prior (explainable) structure if available.
}
\label{fig:fb1}
\end{center}
 \vskip -0.2in
\end{figure}

This paper suggests to decompose convolutional filters in CNN 
into a  truncated expansion with pre-fixed bases in the spatial domain, 
namely the Decomposed Convolutional Filters network (DCFNet),
where the expansion coefficients remain learned from data. 
By representing the filters in terms of functional bases, 
which can come from prior data or task knowledge,
rather than as pixel values, 
the number of trainable parameters is reduced to the expansion coefficients;
and furthermore, 
regularity conditions can be imposed on the filters via the truncated expansion.
For image classification tasks,
we empirically observe that
DCFNet is able to maintain the accuracy 
with a significant reduction in the number of parameters.
 Such observation holds even when random bases are used.

In particular, we adopt in DCFNet the leading Fourier-Bessel (FB) bases \cite{abramowitz1964handbook}, 
which correspond to the low-frequency components in the input.
We experimentally observe the superior performance
of DCFNet with FB bases (DCF-FB)
in both image classification and denoising tasks.
DCF-FB network reduces the response to the high-frequency components in the input,
which are least stable under image variations such as deformation
and often do not affect recognition after being suppressed. 
Such an intuition is further supported by a mathematical analysis of the CNN representation,
where we firstly develop a general result for the CNN representation stability
when the input image undergoes a deformation,
under proper boundedness conditions of the convolutional filters 
(Propositions \ref{prop:l2stable}, \ref{prop:deform1}, \ref{prop:deform2}).
After imposing the DCF structure, 
we show that as long as 
the trainable expansion coefficients at each layer of a DCF-FB network satisfy a boundedness condition,
the $L$-th-layer output is stable with respect to input deformation
and the difference is bounded by the magnitude of the distortion (Theorems \ref{thm:deform3}, \ref{thm:deform4}). 

Apart from FB bases, 
the DCFNet structure studied in this paper is compatible with general choices of bases,
such as standard Fourier bases, wavelet bases,
random bases and PCA bases.
We numerically test several options in Section \ref{sec:4}.
The stability analysis for DCF-FB networks 
can be extended to other bases choices as well,
based upon the general theory developed for CNN representation and using similar techniques.

Our work is related to recent results on the topics of
the usage of bases in deep networks, 
the model reduction of CNN,
as well as the stability analysis of the deep representation.
We review these connections in Section \ref{sec:1-1}.
Finally, though the current paper focuses on supervised networks 
for classification and recognition applications in image data,
the introduced DCF layers are a generic concept and 
can potentially be used in reconstruction and generative models as well. 
We discuss possible extensions in the last section.

\subsection{Related works}\label{sec:1-1}

{\bf Deep network with bases and representation stability}.
The usage of bases in deep networks has been previously studied,
including wavelet bases, PCA bases, learned dictionary atoms, etc.
%
%
Wavelets are a powerful tool in signal processing \cite{mallat2008wavelet} 
and have been shown to be the optimal basis for data representation under generic settings \cite{donoho1994ideal}.  
As a pioneering mathematical model of CNN, the 
{\it scattering transform} \cite{mallat2012group,Bruna2013,Sifre2013} 
used pre-fixed weights in the network which are wavelet filters,
and showed that the representation produced by a scattering network
is stable with respect to certain variations in the input. 
The extension of the scattering transform has been studied
in \cite{wiatowski2015deep,wiatowski2017mathematical} 
which includes a larger class of bases used in the network.
%
%
Apart from wavelet,
deep network with PCA bases has been studied in \cite{chan2015pcanet}.
%
%
Making a connection to dictionary learning~\cite{aharon2006rm},  
\cite{papyan2016convolutional} 
studied deep networks in form of a cascade of convolutional sparse coding layers with theoretical analysis. 
%
%
Deep networks with random weights have been studied in \cite{giryes2016deep},
with proved representation stability.
The DCFNet studied in this paper
incorporates structured pre-fixed bases 
combined by  {\it adapted} expansion coefficients learned from data in a supervised way,
and demonstrates comparable and even improved classification accuracy 
on image datasets.
While the combination of fixed bases and learned coefficients
has been studied in classical signal processing \cite{freeman1991design,mahalanobis1987minimum}, 
dictionary learning \cite{rubinstein2010double} 
and computer vision \cite{henriques2013beyond, bertinetto2016staple}, 
they were not designed with deep architectures in mind.
Meanwhile, the representation stability of DCFNet is inherited thanks 
to the filter regularity imposed by the truncated bases decomposition.

{\bf Network redundancy}.
Various approaches have been studied to suppress redundancy in the weights of trained CNNs, 
including model compression and sparse connections. 
In model compression, network pruning has been studied in \cite{han2015learning}
and combined with quantization and Huffman encoding in \cite{han2015deep_compression}. 
\cite{chen2015compressing} used hash functions to reduce model size without sacrificing generalization performance.
Low-rank compression of filters in CNN has been studied in \cite{DentonZBLF14,ioannou2015training}.
\cite{SqueezeNet, Lin2014} explored model compression with specific CNN architectures, e.g., replacing regular filters with $1 \times 1$ filters.
Sparse connections in CNNs have been recently studied in \cite{ioannou2016deep, anwar2017structured, changpinyo2017power}.
On the theoretical side,
 \cite{bolcskei2017optimal} showed that a sparsely-connected network 
can achieve certain asymptotic statistical optimality. 
The proposed DCFNet relates model redundancy compression to
the regularity conditions imposed on the filters.
In DCF-FB network,
 redundancy reduction is achieved by suppressing
 network response to the high-frequency components in the inputs.

\begin{figure*}[t]
\vskip -0.05in
\begin{center}
 \includegraphics[width= \linewidth, height=0.275\linewidth]{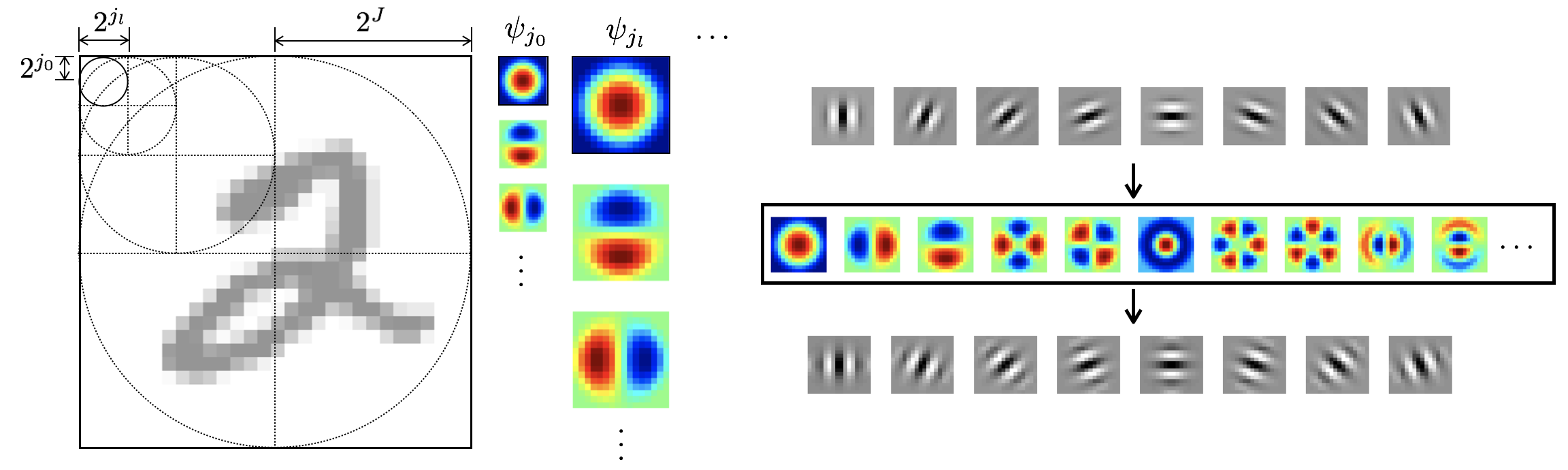} 
 \vskip -0.025 in
\caption{
(Left) Multi-scale convolutional filters and Fourier-Bessel bases in various scales, $j_0 \le \cdots \le j_l \cdots \le J$.
(Right) 
$L\times L$ Gabor filters in 8 directions in size of, $L=11$,
and the approximation by $K$ leading FB bases with a reduction rate of $\frac{K}{L^2} = \frac{1}{3}$.
The truncation incurs almost no change to the filters.
The leading FB bases are shown in the middle panel.
Images rescaled for illustration purpose.	
}
\label{fig:fb2}
\end{center}
\vskip -0.1in
\end{figure*}

%
%
\section{Decomposed Convolutional Filters}\label{sec:2}

\subsection{Notations of CNN}

The output at the $l$-th layer of a convolutional neural network (CNN) can be written as $\{ x^{(l)}(u,\lambda) \}_{u \in \R^2, \lambda \in [M_l]}$,
where $M_l$ is the number of channels in that layer and $[M] = \{1,\cdots, M\}$ for any integer $M$.
A CNN with $L$ layers can be written as a mapping from 
$\{ x^{(0)}(u,\lambda) \}_{u \in \R^2, \lambda \in [M_0]}$
to
$\{ x^{(L)}(u,\lambda) \}_{u \in \R^2, \lambda \in [M_L]}$,
recursively defined via 
$x^{(l)}(u, \lambda) = \sigma( x^{(l)}_{\frac{1}{2}}(u, \lambda) + b^{(l)}(\lambda))$, $\sigma$ being the nonlinear mapping, e.g., ReLU,
and
\begin{equation}\label{eq:conv1}
x^{(l)}_{\frac{1}{2}}(u, \lambda) = \sum_{\lambda' =1}^{M_{l-1}} \int W^{(l)}_{\lambda',\lambda}(v') x^{(l-1)}(u+v', \lambda') dv'.
\end{equation}
The filters $W^{(l)}_{\lambda',\lambda}(u)$ and the biases $b^{(l)}$ are the parameters of the CNN.
In practice, both $x^{(l)}(u,\lambda)$ and $W^{(l)}_{\lambda',\lambda}(u)$ are discretized on a Cartesian grid,
and the continuous convolution in \eqref{eq:conv1} is approximated by its discrete analogue.
Throughout the paper we use the continuous spatial variable $u$ for simplicity.
Very importantly, the filters $W^{(l)}_{\lambda',\lambda}(u)$ are locally supported, 
e.g., on $3 \times 3$ or $5 \times 5$ image patches.

\subsection{Decomposition of convolutional filters}

CNNs typically represent and store filters as vectors of the size of the local patches, 
which is equivalent to expanding the filters under the {\it delta bases}.
Delta bases are not optimal for representing smooth functions. 
For example, 
regular functions have fast decaying coefficients under Fourier bases, 
and natural images have sparse representation under wavelet bases.
DCF layers represent the convolutional filters as a truncated expansion
under basis functions
which are {\it non-adapted} through the training process,
while adaption comes via the combination of such bases.
Specifically, suppose that the convolutional filters $W_{\lambda',\lambda}(u)$ at certain layer,
after a proper rescaling of the spatial variable (detailed in Section \ref{sec:3}), 
are supported on the unit disk $D$ in $\R^2$. 
Given a bases $\{\psi_k\}_k$ of the space $L^2(D)$, 
the filters can be represented as
 \begin{equation}\label{eq:dcf1}
 W_{\lambda',\lambda}(u) = \sum_{k=1}^K (a_{\lambda', \lambda })_k \psi_k(u),
 \end{equation}
 where $K$ is the truncation.
 The decomposition \eqref{eq:dcf1} is illustrated in Figure \ref{fig:fb1},
 and conceptually, it can be viewed as a two-step scheme of a convolutional layer:

\begin{enumerate}

\item
($\Psi$-step) the input is convolved with each of the basis $\psi_k$, $k =1,\cdots, K$, which are {\it pre-fixed}. 
The convolution for each input channel is independent from other channels, 
adding computational efficiency.

\item
($a$-step) the intermediate output is linearly transformed by an effectively fully-connected weight matrix $(a_{\lambda',\lambda})_k$
mapping from index $(\lambda',k)$ to $\lambda$, which is {\it adapted} to data. 
\end{enumerate}

In \eqref{eq:dcf1}, $\psi_k$ can be any bases, 
and we numerically test on different choices in Section \ref{sec:4},
including data-adapted bases and random bases.
All experiments consistently show that the convolutional layers can be drastically decomposed and compressed
with almost no reduction on the classification accuracy, 
and sometimes even using random bases gives strong performance.
In particular, motivated by classical results of harmonic analysis, 
we use FB bases in DCFNet,
with which the regularity of the filters $ W_{\lambda',\lambda}$ can be imposed 
though constraining the magnitude the coefficients $\{(a_{\lambda', \lambda })_k\}_k$ (Proposition \ref{prop:fb2}).
As an example, Gabor filters approximated using the leading FB bases are plotted in the right of Figure \ref{fig:fb2}.
In experiments, DCFNet with FB bases shows superior performance in image classification and denoising tasks 
compared to original CNN and other bases being tested (Section \ref{sec:4}).
Theoretically, 
Section \ref{sec:3} analyzes the representation stability of DCFNet  with respect to input variations,
which provides a theoretical explanation of the advantage of FB bases.

\subsection{Parameter and computation reduction}

Suppose that the original convolutional layer is of size $L \times L \times M' \times M$,
as shown in Figure \ref{fig:fb1}, 
where typically $L= 3$, 5 and usually less than 11,
$M'$ and $M$ grow from $3$ (number of input channels) to a few hundreds in the deep layers in CNN.
After switching to the DCFNet as in \eqref{eq:dcf1}, 
there are $M' \times M \times K$ tunable parameters $(a_{\lambda', \lambda })_k$.
Thus the number of parameters in that layer is a factor $\frac{K}{L^2}$ smaller,
which can be significant if $K$ is allowed to be small, particularly when $M'$ and $M$ are large.

The theoretical computational complexity can be calculated directly.
Suppose that the input and output activation is $W \times W$ in spatial size,
the original convolutional layer needs $M' W^2 \cdot M(1+2 L^2)$ flops
(the number of convolution operations is $M'M$, 
each take $2L^2W^2$  flops, and the summation over channels take an extra $W^2 M'M$).
In contract, a DCF layer takes  $M'W^2\cdot 2K(L^2+M)$ flops,
($M'K$ many convolutions in the $\Psi$ step, and $2KM'MW^2$ flops in the $a$ step).
Thus when $M \gg L^2$,
the leading computation cost is $\frac{K}{L^2}$ of that of a regular CNN layer.

The reduction rate of $\frac{K}{L^2}$ in both model complexity and theoretical computational flops 
is confirmed on actual networks used in experiments, c.f. Table \ref{tab:dcf-acc}.

%
%
\section{Analysis of Representation Stability}\label{sec:3}

The analysis in this section is firstly done for regular CNN and then the conditions on filters are reduced to generic conditions on learnt coefficients in a DCF Net.
In the latter, the proof is for the Fourier-Bessel (FB) bases, and can be extended to other bases using  similar techniques.

\subsection{Stable representation by CNN}

We consider the spatial deformation operator denoted by $D_{\tau}$, 
where $\tau:\mathbb{R}^{2}\to\mathbb{R}^{2}$ and is $C^{2}$, $\rho(u)=u-\tau(u)$, and 
\[
D_{\tau}x(u,\lambda)=x(\rho(u),\lambda),\quad\forall u,\lambda.
\]
We assume that the distortion is controlled, 
and specifically,

\begin{itemize}
\item[ ]
{\bf (A0) } $|\nabla\tau|_{\infty} = \sup_{u} \| \nabla \tau(u) \| <\frac{1}{5}$,
$\|\cdot\|$ being the operator norm.
\end{itemize}

The choice of the constant $\frac{1}{5}$ is purely technical.
Thus $\rho^{-1}$ exists, at least locally. 
Our goal is to control $\|x^{(L)}[D_\tau x^{(0)}] -x^{(L)}[x^{(0)}]\|$,
namely when the input undergoes a deformation the output at $L$-the layer is not severely changed.
We achieve this in two steps: 
(1)
 We show that $ \| D_{\tau}x^{(L)}[x^{(0)}] - x^{(L)}[D_{\tau}x^{(0)}] \| $
is bounded by the magnitude of deformation up to a constant proportional to the norm of the signal,
c.f. Proposition \ref{prop:deform1}.
(2)
We show that $x^{(L)}$ is stable under $D_\tau$ when $L$ is large,
 c.f. Proposition \ref{prop:deform2}. 
To proceed, define the $L^2$ norm of $x(u,\lambda)$
to be
\begin{equation}
\label{eq:normxl2def}
\|x\|^{2}= \frac{1}{M}\sum_{\lambda\in[M]} \frac{1}{|\Omega|}  \int_{\R^2} |x(u,\lambda)|^{2}du,
\end{equation}
where $|\Omega|^2 = (2\cdot 2^J)^2$ is the area of the image-support domain, c.f. Figure \ref{fig:fb2}.
We assume that 
\begin{itemize}
\item[ ]
{\bf (A1)} $\sigma: \R \to \R$ is non-expansive, 
\end{itemize}
which holds for ReLU. We also define the constants
\begin{align}
B_l & : = \max \{ 
	\sup_{\lambda} \sum_{\lambda'=1}^{M_{l-1}}  \| W^{(l)}_{\lambda', \lambda}\|_1, 
	\sup_{\lambda'}  \frac{M_{l-1}}{M_l}  \sum_{\lambda=1}^{M_{l}}  \| W^{(l)}_{\lambda', \lambda}\|_1 \},  \nonumber \\
C_l & : = \max \{ 
	\sup_{\lambda} \sum_{\lambda'=1}^{M_{l-1}}  \| |v || \nabla W^{(l)}_{\lambda', \lambda}(v) |\|_1, \nonumber \\
      & ~~~~~~~~ \sup_{\lambda'}  \frac{M_{l-1}}{M_l}  \sum_{\lambda=1}^{M_{l}}  \| |v || \nabla W^{(l)}_{\lambda', \lambda}(v) |\|_1 \},
      \label{eq:BlCldef}
\end{align}
where $\| |v| |\nabla W(v)| \|_1$ denotes $\int_{\R^2} |v| |\nabla W(v)|  dv$.

Firstly, the following proposition shows that the layer-wise mapping is non-expansive whenever $B_l \le 1$, 
the proof of which is left to Supplementary Material (S.M.).
\begin{proposition}\label{prop:l2stable}
In a CNN, under (A1), if $B_l \le 1$ for all $l$, 

(a) 
The mapping of the $l$-th convolutional layer (including $\sigma$), denoted as $x^{(l)}[x^{(l-1)}]$, is non-expansive,
i.e., $ \| x^{(l)}[ x_1 ] -  x^{(l)}[ x_2 ] \| \le \| x_1 - x_2\| $ for arbitrary $x_1$ and $x_2$. 

(b) $\| x_c^{(l)} \| \le \| x_c^{(l-1)} \|$ for all $l$, where $x_c^{(l)}(u,\lambda) = x^{(l)}(u,\lambda) - x_0^{(l)}(\lambda)$ is the centered version of $x^{(l)}$, 
$x_0^{(l)}$ being the output at the $l$-th layer from a zero input at the bottom layer.
 As a result,
 $\| x_c^{(l)} \|  \le \|x_c^{(0)}\| = \|x^{(0)}\|$.
\end{proposition}

To switch the operator $D_\tau$ with the $L$-layer mapping $x^{(L)}[x^{(0)}]$,
the idea is to control the residual of the switching at each layer, which is the following lemma proved in S.M..

\begin{lemma}\label{lemma:commuting}
In a CNN, under (A0) (A1), $B_l, C_l$ as in \eqref{eq:BlCldef},
\begin{align*}
\|D_{\tau} x^{(l)}[ x^{(l-1)} ] 
& - x^{(l)}[D_{\tau} x^{(l-1)}] \| \\
& \le 4 (  B_l + C_l )
    \cdot|\nabla \tau|_{\infty} \| x_c^{(l-1)}\|,
\end{align*}
where $x_c^{(l)}$ is as in Proposition \ref{prop:l2stable}.
 \end{lemma}

We thus impose the assumption on the filters to be
\begin{itemize}
\item[ ]
{\bf (A2)} For all $l$, $B_l$ and $C_l$ as in \eqref{eq:BlCldef} are less than 1.
\end{itemize}

The assumption (A2) corresponds to a proper scaling of the convolutional filters 
so that the mapping in each convolutional layer 
 is non-expansive (Proposition \ref{prop:l2stable}),
 and in practice, this can be qualitatively maintained 
 by the standard normalization layers in CNN. 

Now we can bound the residual of a $L$-layer switching to be additive as $L$ increases:

\begin{proposition}\label{prop:deform1}
In a CNN, under (A0), (A1), (A2), 
\begin{equation}\label{eq:thmdeform1}
\| D_{\tau}x^{(L)}[x^{(0)}] 
 - x^{(L)}[D_{\tau}x^{(0)}] \| 
 \le  8 L | \nabla \tau |_\infty \|x^{(0)}\|.
\end{equation}
\end{proposition}

Proof is left to S.M.
We remark that it is possible to derive a more technical bound in terms of the constants $B_l$, $C_l$ 
without assuming (A2), using the same technique.
We present the simplified result here.

In the later analysis of DCF Net, (A2) will be implied by
a single condition on the bases expansion coefficients, c.f. (A2').

To be able to control $\|D_\tau x^{(L)} - x^{(L)}\|$, we have the following proposition, 
proved in S.M.
\begin{proposition}\label{prop:deform2}
In a CNN, under (A1),
\[ 
\| D_\tau x^{(l)} - x^{(l)} \| \le 2 |\tau|_\infty D_l  \|x_c^{(l-1)}\|,
\]
where 
$x_c^{(l)}$ is
as in Proposition \ref{prop:l2stable}, 
and
$D_l : =  \max \{ 
	\sup_{\lambda} \sum_{\lambda'=1}^{M_{l-1}}  \| \nabla W^{(l)}_{\lambda',\lambda}\|_1,
         \sup_{\lambda'}  \frac{M_{l-1}}{M_l}  \sum_{\lambda=1}^{M_{l}}  \| \nabla W^{(l)}_{\lambda',\lambda}\|_1 \}$.

\end{proposition}

One may notice that $|\tau|_\infty$ is not proportional to $|\nabla \tau|_\infty$ when the deformation happens on a 
large domain, e.g., a rotation.
It turns out that the multi-scale architecture of CNN 
induces a decrease of the quantity $D_l$ proportional to the inverse of the domain diameter,
which compensate the increase of $|\tau|_\infty$ as scale grows, 
as long as the rescaled filters are properly bounded in integral.
Thus a unified deformation theory can be derived for DCFNets, see next section.

\begin{figure*}[t]
\begin{center}
 \includegraphics[width=0.85\linewidth, height = 0.275\linewidth]{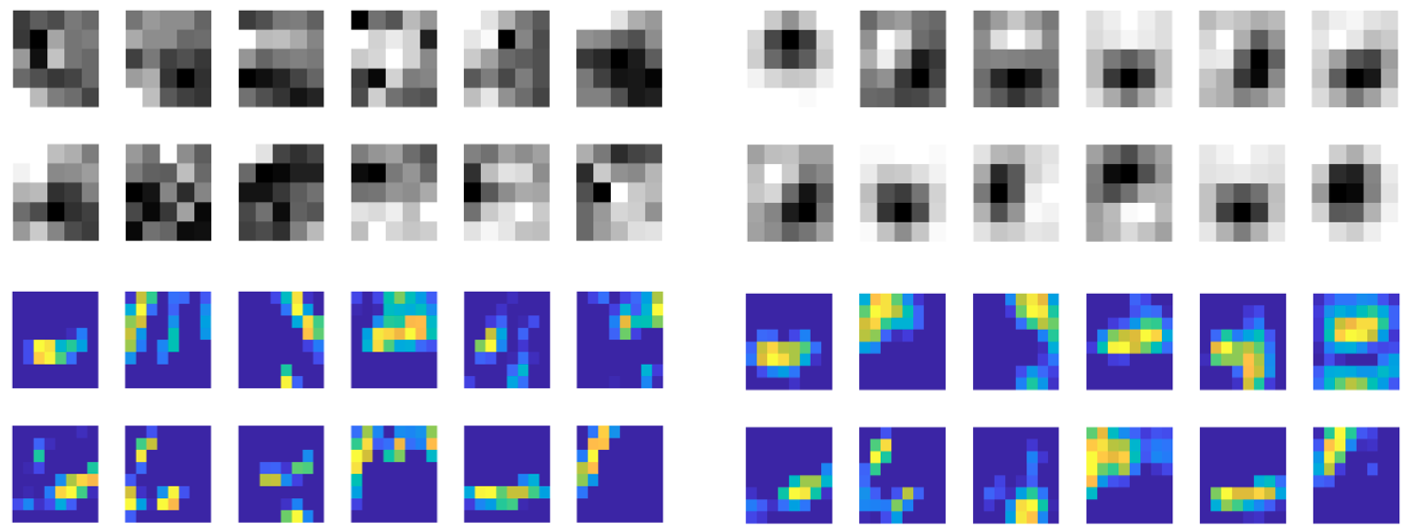} 
\caption{
Example convolutional filters (upper)
and network outputs (bottom)
in the second layer of 
 a Conv-2 net  trained on MNIST (left)
and
 the corresponding DCFNet using 3 FB bases (right).
The filters in DCFNet are visibly smoother than those in the CNN,
so are the network outputs.
Classification accuracy of the two networks is comparable, c.f. Table \ref{tab:dcf-acc}. 
}
\label{fig:filter1}
\end{center}
\vskip -0.125 in
\end{figure*}

\subsection{Multi-scale filters and Fourier Bessel (FB) bases}

Due to the downsampling (``pooling") in CNN,
the support of the $l$-th layer filters $W^{(l)}_{\lambda',\lambda}$ enlarges as $l$ increases.
Suppose that the input is supported on $\Omega$ which is a $ (2\cdot2^J)\times (2\cdot2^J)$ domain,
and the CNN has $L$ layers. 
In accordance with the $2\times 2$ pooling, 
we assume that $ W^{(l)}_{\lambda',\lambda}$ is supported on $D(j_l)$, 
vanishing on the boundary, 
where $D(j)$ is a disk of radius $2^{j}$,
$j_0 \le \cdots \le j_L \le J$,
and $D(j_0)$ is of size of patches at the smallest scale. 
Let $\{\psi_k\}_k$ be a set of bases supported on the unit disk $D(0)$,
and we introduce the rescaled bases 
\[
\psi_{j,k}(u) = 2^{-2j} \psi_k (2^{-j} u), \quad u\in D(j),
\]
where the normalization $2^{-2j}$ is introduced so that $\| \psi_{j,k}\|_1 = \| \psi_{k}\|_1$, 
where $\| f \|_1 : = \int_{\R^2} |f(u)| du$.
The multiscale filters and bases are illustrated in the left of Figure \ref{fig:fb2}.
By \eqref{eq:dcf1}, we have that
 \begin{equation}
 \label{eq:dcf2}
 W^{(l)}_{\lambda',\lambda}(u) = \sum_{k} (a^{(l)}_{\lambda', \lambda })_k \psi_{j_l,k}(u),
 \quad u\in D(j_l).
 \end{equation}

While DCFNet is compatible with general choices of bases, 
we focus on the FB bases in this section as an example.
 FB bases $\psi_k$ are indexed by $k = (m,q)$ where $m$ and $q$ are the angular and radial frequencies respectively. 
 They are supported on the unit disk $D =D(0)$, and in polar coordinates, 
 \[
 \psi_{m,q} (r, \theta) = c_{m,q} J_m (R_{m,q} r) e^{i m \theta}, 
 \, r \in [0,1], \, \theta \in [0,2\pi],
 \]
 where $J_m$ is the Bessel function of the first kind, 
 $m$ are integers, $q=1,2,\cdots$,
 $R_{m,q}$ is the $q$-th root of $J_m$,
 and $c_{m,q}$ is the normalizing constant s.t. 
 $\langle \psi_{m,q}, \psi_{m',q'} \rangle = \int_D \psi_{m,q}(u) \psi_{m',q'}^* (u) du = \pi \delta_{m,m'}\delta_{q,q'}$.
 Furthermore, FB bases are eigenfunctions of the Dirichlet Laplacian on $D$, 
 i.e.,
  $-\triangle \psi_k = \mu_k \psi_k$, where $\mu_{m,q} = R_{m,q}^2$. 
  The eigenvalue
  $\mu_k$ 
grows as $k$ increases (Weyl's law). 
  Thus FB bases can be ordered by $k$ so that $\mu_k$ increases,
 of which the leading few are shown in Table \ref{tab:fb2} and illustrated in Fig. \ref{fig:fb2}.
 In principle, the frequency $q$ and $m$ should be truncated according to the Nyquist sampling rate.
 This truncation turned out to be not often used in our setting,
  due to the significant bases truncation in DCFNet.

\begin{table}[t]
\scriptsize
\begin{centering}
\begin{tabular}[t]{ c | c  c  c  c  c  c  c  c }
\hline 
$k$ 		&	1  	&	2,3 	&	4,5	&	 6	&	7,8	&	9,10	&	11,12 &	13,14 	\\
\hline	
$m$		&	0    	&	1      & 	2	&	0   & 		3	&   	1	&	4	&	 2		\\
$q$		&	1	&	1	&	1	&	2   &	   	1   	&	2	&	1	&	 2		\\
$\mu_k$  &	5.78  &	14.68 &	26.37 &	30.47 &	40.71 &	49.22  &	57.58 &	 70.85	\\
\hline         
\end{tabular}
\caption{\label{tab:fb2}
The angular frequency $m$,
radial frequency $q$
and Dirichlet eigenvalue $\mu_k$
of  
the first $14$ Fourier-Bessel bases.
Two $k$ corresponds to one pair of $(m,q)$ when $m\neq 0$
due to that both real and complex parts of the bases are used as real-valued bases.
}
\vskip -0.1in
\end{centering}
\end{table}

 The key technical quantities in the stability analysis of CNN are 
$\| W^{(l)}_{\lambda',\lambda} \|_1$ and 
$ \| |v| |\nabla W^{(l)}_{\lambda',\lambda} (v)| \|_1$,
and with FB bases,  
these integrals are bounded by a 
$\mu_k$-weighted $L^2$-norm of $a^{(l)}_{\lambda',\lambda}$ 
defined as 
$ \| a \|_{FB} =   (\sum_k \mu_k a_k^2 )^{1/2}$
for all $l$. The following lemma and proposition are proved in S.M.

\begin{lemma}\label{lemma:fb1}
Suppose  that $\{ \psi_k \}$ are FB bases, the function $F(u) = \sum_k a_k \psi_k(u)$ is smooth on the unit disk.
Then $\frac{1}{\sqrt{\pi}}\| \nabla F\|_2 =  \| a \|_{FB} $, 
where $\mu_k$ are the eigenvalues of $\psi_k$ as eigenfunctions of the negative Dirichlet laplacian on the unit disk. 
As a result, $\| \nabla F\|_1 \le \pi \| a \|_{FB}$.
\end{lemma}

\begin{proposition}\label{prop:fb2}
Using FB bases,  $\| |v| |\nabla W^{(l)}_{\lambda',\lambda} (v)| \|_1$ and 
$\| W^{(l)}_{\lambda',\lambda} \|_1$ are bounded by $\pi \|a^{(l)}_{\lambda',\lambda}\|_{FB}$ for all $\lambda', \lambda$ and $l$.
\end{proposition}

Notice that the boundedness of $\| a \|_{FB}$ implies a decay of $|a_k|$ at least as fast as $\mu_k^{-1/2}$.
This justifies the truncation of the FB expansion to the leading few bases, 
which correspond to the low-frequency modes.

Proposition \ref{prop:fb2} implies that $B_l$ and $C_l$ are all bounded by $A_l$ defined as
\[
\begin{split}
A_l & : = \pi \max \{ 
	 \sup_{\lambda} \sum_{\lambda'=1}^{M_{l-1}}  \| a^{(l)}_{\lambda', \lambda}\|_{FB} ,  \\
      & ~~~~~~~~ \sup_{\lambda'}  \frac{M_{l-1}}{M_l}  \sum_{\lambda=1}^{M_{l}}  \| a^{(l)}_{\lambda', \lambda}\|_{FB} \}.
\end{split}	
\]
Then we introduce
\begin{itemize}
\item[ ]
{\bf (A2')} 
For all $l$, $A_l \le 1$,
\end{itemize}
and the result of Proposition \ref{prop:deform1} extends to DCFNet:

\begin{theorem}\label{thm:deform3}
In a DCFNet with FB bases,
under (A0),(A1), (A2'), then
\begin{equation*}
\|D_\tau x^{(L)}[x^{(0)}] - x^{(L)}[D_\tau x^{(0)}]  \| \le 
8 L | \nabla \tau |_\infty \|x^{(0)}\|.
\end{equation*}
\end{theorem}

Combined with Proposition \ref{prop:deform2},
 we have the following deformation stability bound, proved in S.M.:

\begin{theorem}\label{thm:deform4}
In a DCFNet with FB bases, 
under (A0),(A1), (A2'), 
\begin{equation}
\label{eq:thm-deform4}
\begin{split}
\| x^{(L)}[x^{(0)}] 
& - x^{(L)}[D_\tau x^{(0)}]  \|  \\
& \le 
(  8 L | \nabla \tau |_\infty   + 2 \cdot 2^{- j_L} |\tau|_\infty   ) \|x^{(0)}\|.
\end{split}
\end{equation}
\end{theorem}

%
%
\section{Experiments}\label{sec:4}

\begin{figure*} [t]
\vskip -0.1in
\centering
 \subfloat[Original] {\label{fig:original} \includegraphics[angle=0, width=.26\textwidth]{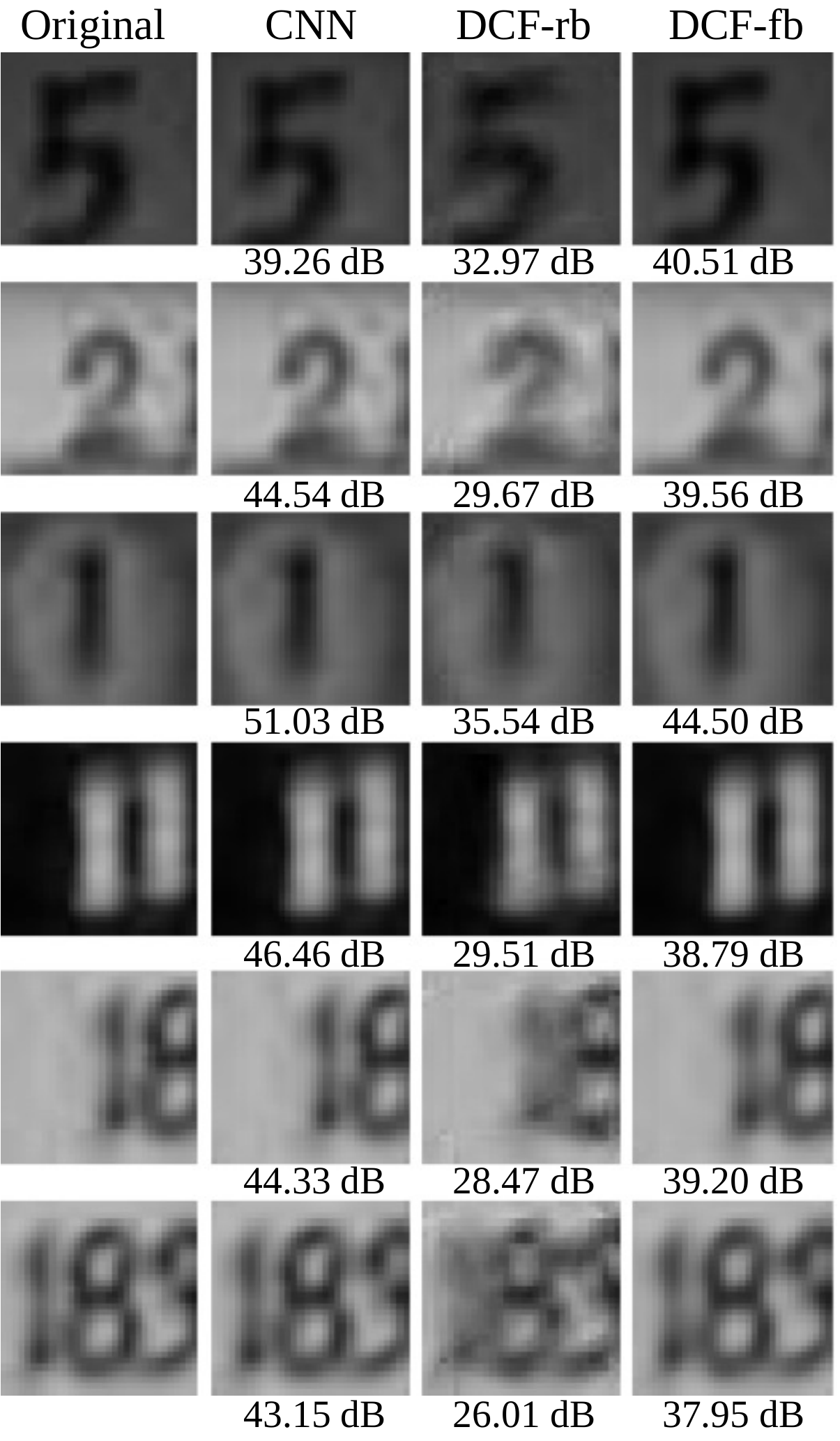} \hspace{7pt}}
 \subfloat[Gaussian noise] {\label{fig:gaussian} \includegraphics[angle=0, width=.26\textwidth]{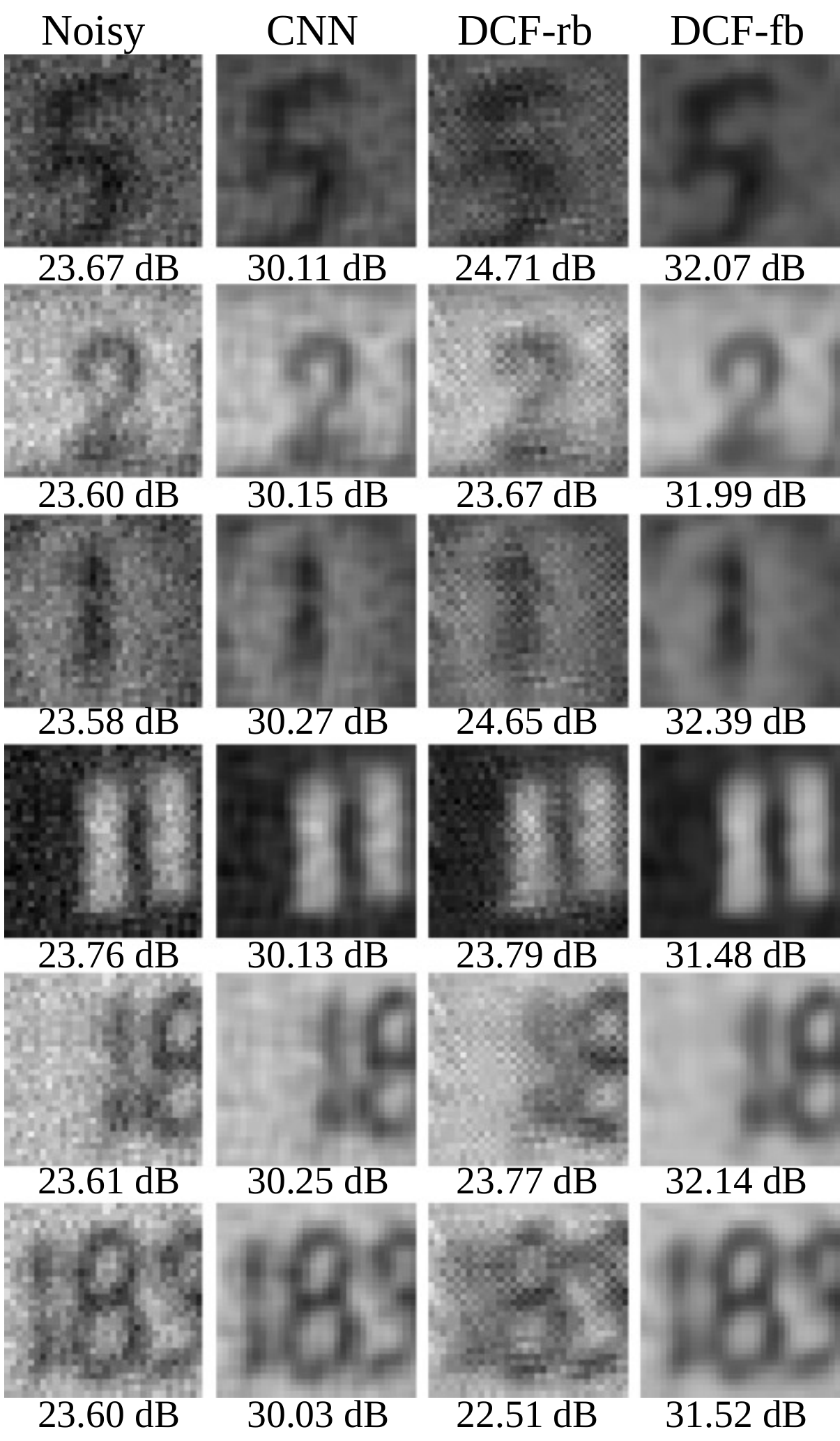} \hspace{7pt}}
    \subfloat[Speckle noise] {\label{fig:speckle} \includegraphics[angle=0,  width=.26\textwidth]{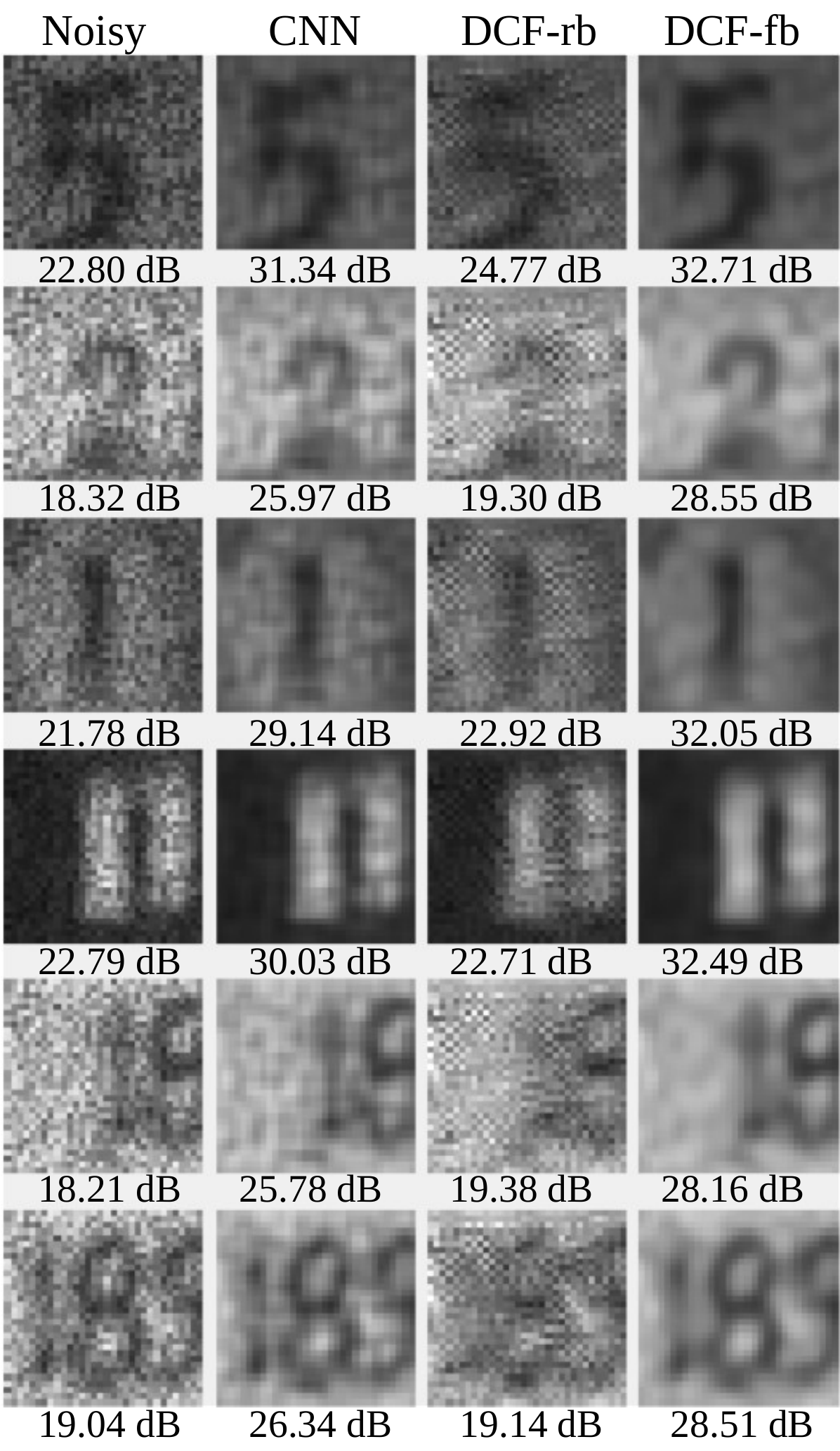} \hspace{0pt}}
\caption{
Examples (randomly selected) of image denoising on the SVHN dataset with PSNR values shown. 
The average PSNR over the entire test set including 26,032 samples:
with Gaussian noise, 30.01 for CNN, 31.24 for DCF-fb;
with Speckle noise, 28.15 for CNN, 29.84 for DCF-fb.
}
\vskip -0.05in
\label{fig:denoising}
\end{figure*}

\begin{table}[b]
\vskip -0.1 in
\small
\begin{centering}
\begin{tabular}[t]{  l | l }
\hline
~~Conv-2 &   	 ~~Conv-3          \\
\hline
c5x5x1x16 ReLu mp3x3& c5x5x3x64  ReLu  mp3x3\\
c5x5x16x64  ReLu  mp3x3 & 	 c5x5x64x128  ReLu  mp3x3\\
  fc128  ReLu  fc10 & 	 c5x5x128x256  ReLu  mp3x3\\
			&	 fc512  ReLu  fc10 \\
\hline       	
\end{tabular}
\caption{\label{tab:network-arch-1}
\small
CNN network architectures used in MNIST, SVHN, and CIFAR10 experiments. 
c$L$x$L$x$M'$x$M$ stands for a convolutional layer of patch size $L$x$L$ and input (output) channel $M'$ ($M$).
mp$L$x$L$ stands for $L$x$L$ max-pooling. 
For the corresponding DCFNets, each $L$x$L$x$M'$x$M$ CNN conv layer 
is expended over $K$ $L \times L$ bases for trainable coefficients implemented as a $1 \times 1 \times M'K \times M$ conv layer.
}
\end{centering}
\vskip 0.1 in
\end{table}

In this section, we experimentally demonstrate that convolutional filters in CNN can be decomposed as a truncated expansion with pre-fixed bases, where the expansion coefficients remain learned from data.
Though the number of trainable parameters are significantly reduced,
the accuracy in tasks such as image classification and face verification
is still maintained.
Such empirical observations hold for data-independent Fourier-Bessel (FB) and random bases, 
and data-dependent PCA bases.

\subsection{Datasets}

We perform an experimental evaluation on DCFNets using the following public datasets:

{\bf MNIST.} $28 \times 28$ grayscale images of digits from 0 to 9,
 with 60,000 training and 10,000 testing samples. 

{\bf SVHN.} The Street View House Numbers (SVHN)
dataset \cite{svhn} contains $32 \times 32$ colored images of digits 0 to 9, 
 with 73,257  training and 26,032 testing samples.
The additional training images were not used. 

{\bf CIFAR10.} The dataset \cite{cifar} 
contains $32 \times 32$ colored images from 10 object classes,
with 50,000  training and 10,000 testing samples. 

{\bf VGG-Face.} A large-scale face dataset, 
which contains about 2.6M face images from over 2.6K people \cite{vgg-face}.
\footnote{The software is publicly available at \url{https://github.com/xycheng/DCFNet}.}

\subsection{Object classification} 

In our object classification experiments,
we evaluate the DCFNet with three types of predefined bases: Fourier-Bessel bases (DCF-FB), 
random bases which are generated by Gaussian vectors (DCF-RB), 
and PCA bases which are principal components of the convolutional filters in a pre-trained corresponding CNN model (DCF-PCA).

Three CNN network architectures are used for classification, Conv-2 and Conv-3 shown in Table~\ref{tab:network-arch-1}, and VGG-16 \cite{simonyan2014very}.
To generate the corresponding 
DCFNet structure from CNN, each CNN conv layer is expended over a set of pre-defined bases, 
and the obtained trainable expansion coefficients are implemented as a $1 \times 1$ conv layer. 
For example, a $5 \times 5 \times M' \times M$ conv layer is expended over $K$ $5 \times 5$ bases for trainable coefficients in a $1 \times 1 \times M'K \times M$ convolutional layer.
 $K$ denotes the number of basis used, and we evaluate multiple $K$ for different levels of parameter reduction.
 In order to be compatible with existing deep learning frameworks, pre-fixed bases are currently implemented as regular convolutional layers with zero learning rate.
The additional memory cost incurred in such convenient implementation can be eliminated with a more careful implementation, 
as bases are pre-fixed and the addition across channels can be computed on the fly.

The classification accuracy using DCFNets on various datasets are shown in Table \ref{tab:dcf-acc}.
We observe that, by using only 3 Fourier-Bessel (FB) bases, 
we already obtain comparable accuracy as the original full CNN models on all datasets,
while using $12\%$ parameters for $5 \times 5$ filters.
When more FB bases are used, DCFNets outperform corresponding CNN models, still with significantly less parameters.
As FB bases correspond to the low-frequency components in the inputs, 
DCF-FB network
responds less to the high-frequency nuance details, 
which are often irrelevant for classification tasks. 
The superiority of DCF-FB network is further shown with less training data. 
For SVHN with 500 training samples, 
the testing accuracy (on a 50,000 testing set) of regular CNN and DCF-FB are 63.88\% and 66.79\% respectively. 
With 1000 training samples, the test accuracy are 73.53\% v.s. 75.45\%. 
Surprisingly, we observe that DCF with random bases also report acceptable performance.

Both the FB and random bases are data independent. 
For comparison purposes, 
we also evaluate DCFNets with data dependent PCA bases, 
which are principal components of corresponding convolutional filters in pre-trained CNN models.
When the CNN model is pre-trained with all training data, PCA bases (pca-f) shows comparable performance as FB bases. 
However, 
the quality of the PCA bases (pca-s) degenerates, when only a
randomly selected
subset of the training set is used for the pre-training.

\begin{table}[h]
\begin{centering}
\scriptsize
\begin{tabular}[t]{ l | c  c | c c | c c  }
\hline
 \multicolumn{7}{c}{MNIST   conv-2, 5x5 }   \\
\hline
                         &  	fb			&  	rb 	&	pca-s 	& 	pca-f 	   	&  	\# param.	&  \# MFlops  \\
\hline
CNN			&    \multicolumn{4}{|c|}{99.40	}	  				   			& 2.61$\times 10^4$	 & 3.37	\\
\hline
$K$=14		&    99.47 			&	99.35 	&	99.38	&	99.41		& 1.46$\times 10^4$	 & 2.40 	\\
$K$=8		&    99.48	 		&	99.26   	& 	99.28	&	99.45		& 8.40$\times 10^3$	 & 1.37	\\
$K$=5		&    99.39			&	99.28 	& 	99.28	&	99.43		& 5.28$\times 10^3$  & 0.86	\\
$K$=3		&    99.40			&	98.69	&	99.19	&	99.35		& 3.20$\times 10^3$  & 0.51	\\        	
 \hline
     \multicolumn{7}{c}{SVHN  conv-3, 5x5 }   \\
\hline
                         &  fb			& 	rb 		&  	pca-s 	& 	pca-f			&    	\# param.	&  \# MFlops  \\
\hline
CNN			&  \multicolumn{4}{|c|}{94.22}								& 1.03$\times 10^6$	 	&	201.64	\\
\hline
$K$=14		& 94.63 		&	93.75	&	94.52	&	94.42		& 5.74$\times 10^5$		 & 	121.91  \\
$K$=8		& 94.39 		&	92.05	&	93.85	&	94.30		& 3.30$\times 10^5$		 &	69.67	\\
$K$=5		& 93.93		&	91.28	&	92.34	&	94.03		& 2.06$\times 10^5$		 &	43.55	\\
$K$=3		& 92.84		&	88.47	&	91.88	&	93.10		& 1.24$\times 10^5$		 & 	26.13	\\        	
 \hline
     \multicolumn{7}{c}{Cifar10  conv-3, 5x5 }   \\
\hline
                         &  	fb		&  	 rb  		& 	pca-s	 & 	pca-f			&  	\# param.	&  \# MFlops  \\
\hline
CNN			& \multicolumn{4}{|c|}{ 85.66}								&  			&  \\
\hline
$K$=14		& 85.88  		&	84.76	&	85.27	&	85.34		& 		 & 		\\
$K$=8		& 85.30		&	81.27	&	84.70	&	85.09		& \multicolumn{2}{c}{(same as above)}		\\
$K$=5		& 84.35		&	77.96	&	83.12	&	83.94		& 		 &		\\
$K$=3		& 83.12		&	74.05	&	80.94	&	82.91		&  		 & 		\\        	
 \hline
  \multicolumn{7}{c}{ Cifar10  vgg-16, 3x3}   \\
\hline
                         &   	fb		&  	rb  		&  	pca-s 	&  	pca-f		   	&  	\# param.	&  \# MFlops  \\
\hline
CNN			&   \multicolumn{4}{|c|}{87.02}								& 	1.47$\times 10^7$		& 547.20	\\
\hline
$K$=5		&  87.79		&  84.16		&	87.98	&	87.60		& 	8.18$\times 10^6$		& 311.68		\\
$K$=3		&  88.21 		&  78.46		&	87.45	&	87.54		&	4.91$\times 10^6$	 	& 187.02		\\
 \hline
\end{tabular}
\caption{\label{tab:dcf-acc}
Classification accuracy using DCFNets on various image benchmarks with different number of bases $K$. 
``fb" and ``rb" stand for Fourier-Bessel bases and random bases respectively. 
``pca-s" and ``pca-f" stand for PCA bases computed from a network pre-trained on a small subset of training images 
(1,000 random samples) and the full training set respectively.
``\# param." is number of parameters in all convolutional layers,
and MFlops is the number of flops in all convolutional layers (including ReLU).
}
\end{centering}
\vskip -0.2in
\end{table}

\subsection{Image denoising}

To gain intuitions behind the superior classification performance of DCFNet,
 we conduct a set of ``toy" image denoising experiments on the SVHN image dataset.
We take the first three $5\times5$ convolution blocks from the Conv-3 CNN network in Table~\ref{tab:network-arch-1},  
which is used in our SVHN object classification experiments. We remove all pooling layers, 
and append at the end an FC-256 followed with a Euclidean loss layer. 
We then decompose each $5\times5$ conv layer in this CNN network over 3 random bases and 3 FB bases respectively, to produce DCF-RB and DCF-FB networks. 

We use SVHN training images with their gray-scale version as labels to train all three networks to simply reconstruct an input image (in gray-scale). 
Figure~\ref{fig:denoising} shows how three trained networks behave while reconstructing examples from the SVHN testing images.
Without noise added to input images, Figure~\ref{fig:original}, all three networks report decent reconstruction, while DCF-RB shows inferior to both CNN and DCF-FB. PSNR values indicate CNN often produces more precise reconstructions; 
however, those missing high-frequency components in DCF-FB reconstructions are mostly nuance details.
With noise added as in figures ~\ref{fig:gaussian} and \ref{fig:speckle}, DCF-FB produces significantly superior reconstruction over both CNN and DCF-RB, with about one tenth of the parameter number of CNN. 

The above empirical observations clearly indicate that Fourier-Bessel bases, which correspond to the low-frequency components in the inputs, 
enable DCF to ignore the high-frequency nuance details, which are often less stable under input variations, and mostly irrelevant for tasks such as classification. 
 Such empirical observation provides good intuitions behind the superior classification performance of DCF, and is also consistent with the theoretical analysis on representation stability in Section~\ref{sec:3}.

\subsection{Face verification}

We present a further evaluation of DCFNet on face verification tasks using ``very deep" network architectures, which comprise a long sequence of convolutional layers.
In order to train such complex networks, we adopt a very large scale \emph{VGG-face} \cite{vgg-face} dataset, which contains about 2.6M face images from over 2.6K people.

As shown in Table~\ref{tab:facenet}, we adopt the VGG-Very-Deep-16 CNN architecture as detailed in \cite{vgg-face} by modifying layer 32 and 35 to change output features from 4,096 dimension to 512. Such CNN network comprises 16 weight layers, and all except the last Fully-Connected (FC) layer utilize $3\times3$ or $5\times5$ filters.

The input to both CNN and DCFNet are face images of size $224 \times 224$ (with the average face image subtracted). As shown in Table~\ref{tab:vgg-face}, with FB bases,
even only using $\frac{1}{3}$ parameters at weight layers ($K=3$ for $3 \times 3$, $K=8$ for $5 \times 5$), the DCFNet shows similar verification accuracy as the CNN structure on the challenging LFW benchmark. 
Note that our CNN model outperforms the \emph{VGG-face} model in \cite{vgg-face}, and such improvement is mostly due to the smaller output dimension we adopted, as both models share similar architecture and are trained on the same face dataset.

\begin{table}[h!]
\centering
\scriptsize
\label{tab:facenet}
\begin{tabular}{c|c|c}
\hline
Layer & CNN  & DCFNet  \\
\hline					
\multirow{2}{*}{1} &\multirow{2}{*}{conv $3\times 3 \times 3\times 64$} & 3 $3\times 3$ basis\\
					&   & conv $1\times 1 \times 9\times 64$\\									
\hline 
2 & \multicolumn{2}{|c}{ReLu}\\
\hline 
\multirow{2}{*}{3} &\multirow{2}{*}{conv $3\times 3 \times 64 \times 64$} & 3 $3\times 3$ basis\\
					&   & conv $1\times 1 \times 192 \times 64$\\									
\hline 
4-5 & \multicolumn{2}{|c}{ReLu, maxPool  $2\times 2$}\\
\hline 			
\multirow{2}{*}{6} &\multirow{2}{*}{conv $3\times 3 \times 64\times 128$} & 3 $3\times 3$ basis\\
					&   & conv $1\times 1 \times 192\times 128$\\									
\hline 
7 & \multicolumn{2}{|c}{ReLu}\\
\hline 
\multirow{2}{*}{8} &\multirow{2}{*}{conv $3\times 3 \times 128 \times 128$} & 3 $3\times 3$ basis\\
					&   & conv $1\times 1 \times 384 \times 128$\\									
\hline 
9-10 & \multicolumn{2}{|c}{ReLu, maxPool  $2\times 2$}\\			
\hline 						
\multicolumn{3}{c}{(1-31 CNN layers are identical to \emph{vgg-face} model in \cite{vgg-face}.)} \\
\hline 		
\multirow{2}{*}{32} &\multirow{2}{*}{conv $5\times 5 \times 512\times 512$} & 8 $5\times 5$ basis \\
					&   & conv $1\times 1 \times 4096\times 512$\\									
\hline 
33-34 & \multicolumn{2}{|c}{ReLu, dropout}\\
\hline 
\multirow{2}{*}{35} &\multirow{2}{*}{conv $3\times 3 \times 512 \times 512$} & 3 $3\times 3$ basis\\
					&   & conv $1\times 1 \times 1536 \times 512$\\									
\hline 
36-39 & \multicolumn{2}{|c}{ReLu, dropout, FC, softmax}\\
\hline
\end{tabular}
\caption{\label{tab:facenet}
Network architecture for face experiments.
For the corresponding DCFNet, each $L$x$L$x$M'$x$M$ CNN conv layer 
is expended over $K$ $L \times L$ bases for trainable coefficients implemented as a $1 \times 1 \times M'K \times M$ conv layer ($K=3$ for $3 \times 3$, $K=8$ for $5 \times 5$).
}
\end{table}

\begin{table}[h]
\begin{centering}
\small
\begin{tabular}[t]{ c | c  c   c}
 \hline
          &  Accuracy		&  	\# param.	&  \# GFlops  \\
\hline
\emph{VGG-face} & 97.27 \% & - & - \\
\hline
CNN			&    97.65 \%		& 21.26 $\times 10^6$	 & 30.05	\\
DCFNet		&    97.32 \% 			& 7.01 $\times 10^6$	 & 10.09\\
 \hline
\end{tabular}
\caption{\label{tab:vgg-face}
Face verification accuracy on the LFW benchmark.
}
\end{centering}
\end{table}

%
%
\section{Conclusion and Discussion}\label{sec:5}

The paper studies CNNs where the convolutional filters are represented as a truncated expansion under pre-fixed bases
and the expansion coefficients are learned from labeled data. 
Experimentally, we observe that on various object recognition datasets 
the classification accuracy are maintained with a significant reduction of the number of parameters,
and the performance of Fourier-Bessel (FB) bases is constantly superior.
The truncated FB expansion in DCFNet can be viewed as a regularization of the filters.
In other words, 
DCF-FB is less susceptible to the high-frequency components in the input,
which are least stable under expected input variations 
and often do not affect recognition when suppressed. 
This interpretation is supported by image denoising experiments,
where DCF-FB performs preferably over the original CNN and other basis options on noisy inputs.
The stability of DCFNet representation is also proved theoretically,
showing that the perturbation of the deep features 
with respect to input variations can be bounded under generic conditions on the decomposed filters. 

To extend the work, firstly, DCF layers can be incorporated in networks for unsupervised learning, 
for which the denoising experiment serves as a first step.
The stability analysis can be extended by testing the resilience to adversarial noise.
Finally,
more structures may be imposed across the channels,
concurrently with the structures of the filters in space.

\newpage

\bibliographystyle{icml2018}
\bibliography{structnet}

\newpage
\onecolumn

\appendix 

\setcounter{figure}{0}
\renewcommand{\thefigure}{\thesection.\arabic{figure}}

\setcounter{equation}{0}
\renewcommand{\theequation}{\thesection\arabic{equation}}

\section{Proofs}

In the proofs, some technical details are omitted for brevity and readability. 
The full proofs are left to the long version of the work.

\begin{proof}[Proof of Proposition 3.1]
To prove (a), omitting $(l)$ in $W^{(l)}$, and let $M=M_l$, $M' = M_{l-1}$, $B_{\lambda', \lambda} = \| W_{\lambda', \lambda}\|_1$.
By definition of $B_l$, we have that 
\begin{equation}\label{eq:Bl-bound}
\begin{split}
 & \sum_{\lambda' \in [ M' ]}  B_{\lambda', \lambda} \le B_l, \quad \forall \lambda \\
 & \sum_{\lambda \in [ M ]}  B_{\lambda', \lambda} \le B_l \frac{M}{M'}, \quad \forall \lambda'.
\end{split}
\end{equation}
We essentially use Schur's test, being more careful with the summation over $\lambda'$. 
We derive by Cauchy-Schwarz which is equivalent to Schur's test:
\begin{align*}
 & \| x^{(l)}[ x_1 ] - x^{(l)}[ x_2 ]  \|^{2} \cdot |\Omega| M\\
= & \sum_{\lambda\in[M ] }\int
  \left|\sigma ( \sum_{\lambda'\in[ M' ]}\int x_1 (u+v',\lambda')W_{\lambda',\lambda}(v')dv'+b(\lambda) ) 
       - \sigma (\sum_{\lambda'\in[ M' ]}\int x_2(u+v', \lambda')W_{\lambda',\lambda}(v')dv'+b(\lambda) ) \right|^{2}du\\
\le & \sum_{\lambda\in[ M ]}\int\left|
    \sum_{\lambda'\in[ M' ]}
    \int 
     x_1(u+v',\lambda')W_{\lambda',\lambda}(v')dv'-
     \sum_{\lambda'\in[ M' ]}
     \int 
     x_2(u+v', \lambda')W_{\lambda',\lambda}(v')dv'\right|^{2}du\\
= & \sum_{\lambda\in[ M ]}\int\left|
    \sum_{\lambda'\in[ M'  ]}\int(  x_1 - x_2 )(\tilde{v},\lambda')W_{\lambda',\lambda}(\tilde{v}-u)d\tilde{v}
    \right|^{2}du\\
\le & \sum_{\lambda\in[ M ]}
 \int\left(  \sum_{\lambda_{1}'\in[ M' ]}
 \int| (  x_1 - x_2 )
 (  v_{1},\lambda_1'  )|^{2}\left|W_{ \lambda_1' ,\lambda}(v_{1}-u)\right|dv_{1}\right)
 \cdot
   \left(\sum_{\lambda_{2}'\in[ M' ]}
    \| W_{\lambda_2', \lambda}\|_1  
  \right)du\\
\le &  B_l 
 \cdot\sum_{\lambda_{1}'\in[M']}\int|(   x_1 - x_2 )(v_{1},    \lambda_1'   )|^{2}
 \left(
 \sum_{\lambda\in[M]} 
  \| W_{  \lambda_1'  , \lambda}\|_1 
 \right)dv_{1}\\
\le &  
B_l \cdot B_l \frac{M}{M'} \cdot \|  x_1 - x_2 \|^{2} |\Omega| M'  
= 
B_l^2 M  \|  x_1 - x_2 \|^{2} |\Omega|,
\end{align*}
which means that 
\[
 \| x^{(l)}[ x_1 ] - x^{(l)}[ x_2 ]  \| \le B_l \|  x_1 - x_2 \|.
\]
Thus $B_l  \le 1$ implies (a).

To prove (b), 
we firstly verify that $x_0^{(l)}(\lambda)$ indeed is a constant over space for all $\lambda$ and $l$. When $l=0$, $x_0^{(0)}$ is all zero, so the claim is true.
Suppose that the claim holds for $l-1$, then 
\[
x_0^{(l)}(u,\lambda)  = \sigma \left(  
 \sum_{\lambda'} \int x_0^{(l-1)}(\lambda') W^{(l)}_{\lambda',\lambda}(v')dv'  + b^{(l)}(\lambda)
 \right) 
 \]
 which again does not depend on $u$. So we can write  $x_0^{(l)}$ as $x_0^{(l)}(\lambda)$.
 Now by (a),
\[
 \|  x_c^{(l)} \| =  \| x^{(l)}[ x^{(l-1)} ] - x^{(l)}[x_0^{(l-1)}]  \|
\le   \| x^{(l-1)} - x_0^{(l-1)} \|  =  \|  x_c^{(l-1)} \|,
\]
 which proves (b).
\end{proof}
\begin{proof}[Proof of Lemma 3.2]
To illustrate the idea, we first prove the lemma in the one-dimensional case, i.e. $u\in \R$ instead of $\R^2$. We then extend to the 2D case.
In the 1D case, the constant $c_1$ can be improved to be 2, and we only need $|\tau'|_\infty <\frac{1}{2}$. 
In the 2D case, we need $c_1= 4$  as in the final claim.

To simply notation, we denote the mapping $x^{(l)}[x^{(l-1)}]$ as $y[x]$, $x_c^{(l-1)}$ by $x_c$,
$M_{l-1} = M'$, $M_l = M$, and $W^{(l)}$ by $W$.
Let $C_{\lambda',\lambda} = \int |v| |\frac{d}{dv}W_{\lambda',\lambda}(v)| dv$, 
and $B_{\lambda',\lambda} = \int |W_{\lambda',\lambda}(v)|dv$,
then \eqref{eq:Bl-bound} holds, and the same relation holds for $C_{\lambda', \lambda}$ and $C_l$.

 By definition,
\begin{align*}
D_{\tau}y[x](u,\lambda) & =\sigma\left(\sum_{\lambda'\in[M']}\int x(\rho(u)+v',\lambda')W_{\lambda',\lambda}(v')dv'+b(\lambda)\right),\\
y[D_{\tau}x](u,\lambda) & =\sigma\left(\sum_{\lambda'\in[M']}\int x(\rho(u+v'),\lambda')W_{\lambda',\lambda}(v')dv'+b(\lambda)\right).
\end{align*}
Relaxing by removing $\sigma$ as in the proof of Proposition 3.1, one can derive that 
\[
\|D_{\tau}y[x]-y[D_{\tau}x]\|^{2} \cdot |\Omega| M 
\le 
 \|E_{1}+E_{2}\|^{2},
\]
where 
\begin{align*}
E_{1}(u,\lambda) & =\sum_{\lambda'\in[M']}\int x_c(v,\lambda')(W_{\lambda',\lambda}(v-\rho(u))-W_{\lambda',\lambda}(\rho^{-1}(v)-u))dv,\\
E_{2}(u,\lambda) & =\sum_{\lambda'\in[M']}\int x_c(v,\lambda')W_{\lambda',\lambda}(\rho^{-1}(v)-u)(|(\rho^{-1})'(v)|-1)dv.
\end{align*}
Notice that $x$ is replaced by $x_c$ due to the fact that $x$ and $x_c$ differ by a constant field over space for each channel $\lambda'$.
We bound $\|E_{1}\|$ and $\|E_{2}\|$ respectively.

For $E_{1}$, we introduce $k_{\lambda',\lambda}^{(1)}(v,u)=W_{\lambda',\lambda}(v-\rho(u))-W_{\lambda',\lambda}(\rho^{-1}(v)-u)$,
and re-write it as
\[
E_{1}(u,\lambda)=\sum_{\lambda'\in[M']}\int x_c(v,\lambda')k_{\lambda',\lambda}^{(1)}(v,u)dv.
\]
Applying Schur's test as in the proof of Proposition 3.1, one can show that 
\[
\|E_{1}\|  \le 
2 |\tau'|_\infty C_l \sqrt{ M |\Omega| } \|x_c\|
\]
as long as for all $\lambda',\lambda$,
\begin{equation}
\sup_{u}\int\left|k_{\lambda',\lambda}^{(1)}(v,u)\right|dv,\,\sup_{v}\int\left|k_{\lambda',\lambda}^{(1)}(v,u)\right|du
\le 2C_{\lambda',\lambda}|\tau'|_{\infty}.
\label{eq:C-lambda'-lambda}
\end{equation}
 \eqref{eq:C-lambda'-lambda} can be verified by 1D change of variable, and details omitted.

For $E_{2}$, we introduce $k_{\lambda',\lambda}^{(2)}(v,u)=W_{\lambda',\lambda}(\rho^{-1}(v)-u)(|(\rho^{-1})'(v)|-1)$,
and then we have that
\[
\int |k_{\lambda',\lambda}^{(2)}(v,u)|du\le|(\rho^{-1})'(v)-1|\cdot\int|W_{\lambda',\lambda}(u)|du\le2|\tau'|_{\infty}B_{\lambda',\lambda},
\quad\forall v,
\]
where we use $1-(\rho^{-1})'(t)=\frac{-\tau'(\rho^{-1}(t))}{1-\tau'(\rho^{-1}(t))}$
and $|\tau'|<\frac{1}{2}$ to obtain the factor 2. Meanwhile, 
\[
\int|k_{\lambda',\lambda}^{(2)}(v,u)|dv=\int|W_{\lambda',\lambda}(\tilde{v}-u)||1-|\rho'(\tilde{v})||d\tilde{v}\le|\tau'|_{\infty}B_{\lambda',\lambda},
\quad\forall u.
\]
This gives that 
\[
\|E_{2}\| \le 2 |\tau'|_\infty B_l   \sqrt{ M |\Omega| }\|x_c\|.
\]
Putting together we have that \[
\sqrt{ M |\Omega| } \|D_{\tau}y[x]-y[D_{\tau}x]\|
\le \| E_1 + E_2 \|
\le
\|E_{1}\| + \|E_{2}\| 
\le  2 |\tau'|_\infty (C_l + B_l)   \sqrt{ M |\Omega| }\|x_c\| 
\]
which proves the claim in the 1D case.

The extension to the 2D case uses standard elementary techniques. The assumption $| \nabla \tau |_\infty < \frac{1}{5}$ is used to derive that 
$ | |J \rho| - 1 |$, 
$| |J \rho^{-1}| - 1 | \le 4 | \nabla \tau |_\infty$, 
and
$|J \rho|$,
$ |J \rho^{-1}|  \le 2$.
In all the formula, $|(\rho^{-1})'(v)|$ is replaced by the Jacobian determinant $|J \rho^{-1}(v)|$.
The integration in 1D is replaced by that along a segment in the 2D space.
Details omitted.
\end{proof}

\begin{proof}[Proof of Prop. 3.3]
Under these conditions, Proposition 3.1 
applies. Let $c_1 = 4$. 
Introduce the notation
\[
y_{l}=x^{(L)}\circ\cdots\circ D_{\tau}x^{(l)}\circ\cdots\circ x^{(0)},\quad l=0,\cdots,L
\]
where $y_{0}=x^{(L)}[D_{\tau}x^{(0)}]$, and $y_{L}=D_{\tau}x^{(L)}[x^{(0)}]$.
The l.h.s 
equals 
$\|y_{0}-y_{L}\|$,
and we will bound it by $\|y_{L}-y_{0}\|\le \sum_{l=1}^{L}\|y_{l}-y_{l-1}\|$. 
For each $l = 1,\cdots, L$,
\begin{align*}
\|y_{l}-y_{l-1}\|
=  &  \|x^{(L)}\circ\cdots\circ D_{\tau}x^{(l)}\circ x^{(l-1)} \\
    & - x^{(L)}\circ\cdots\circ x^{(l)}\circ D_{\tau}x^{(l-1)}\| \\
\le &  \|D_{\tau}x^{(l)}\circ x^{(l-1)}-x^{(l)}\circ D_{\tau}x^{(l-1)}\| \\
\le & c_1 (C_l + B_l ) |\nabla \tau|_{\infty}\|x_c^{(l-1)}\| \\
\le &  2 c_1  |\nabla \tau|_{\infty}\|x_c^{(l-1)}\| \\
\le & 	 2 c_1 |\nabla \tau|_{\infty}\|x^{(0)}\|,
\end{align*}
where 
the first inequality is by the nonexpansiveness of the \ensuremath{(l+1)} to \ensuremath{L}-th layer,
the second by Lemma 3.2, 
the third by (A2),
and the last by Proposition 3.1 
(b).
Thus,
$\sum_{l=1}^{L}\|y_{l}-y_{l-1}\| \le 2 c_1 L |\nabla \tau|_{\infty}\|x^{(0)}\|$.
\end{proof}

%
%
%
\begin{proof}[Proof of Proposition 3.4]
The technique is similar to that in the proof of Lemma 3.2.
Let the constant on the r.h.s be denoted by $c_2$. 
In the 1D case, the constant $c_2$ can be improved to be 1.
In the 2D case,  $c_2=2$ as in the final claim. 
Details omitted. 
\end{proof}
%
%
%
%
\begin{proof}[Proof of Lemma 3.5]
The first claim is a classical result,
and has a direct proof as $ \int_{D(0)} |\nabla F|^2  = -\int_{D(0)} F \Delta F 
= \langle \sum_k a_k \psi_k, \sum_k a_k \mu_k \psi_k  \rangle
 = \pi \sum_k a_k^2 \mu_k$ by the orthogonality of $\psi_k$, as stated above in the text.
By Cauchy-Schwarz, $\| \nabla F\|_1  \le  \sqrt{\pi} \| \nabla F\|_2 $.
Putting together gives the second claim.
\end{proof}

%
%
%
%
\begin{proof}[Proof of Proposition 3.6]
Omitting $\lambda', \lambda, l$, and let $j_l = j$, we write $W(u) = \sum_k a_k \psi_{j,k}(u)$. Rescaled to $D(0)$, we consider 
$w(u) = \sum_k a_k \psi_k(u)$, and one can verify that $\| |v| |\nabla W(v)| \|_1 = \| |v| |\nabla w(v) | \|_1$, 
and $\| W\|_1 = \|w\|_1$. 
Meanwhile, $ \int_{D(0)} |v| |\nabla w(v)| dv \le \int_{D(0)} |\nabla w(v)| dv$ by  that $|v| \le 1$, 
and $\| w \|_1 \le \| \nabla w\|_1$ by Poincar\'e inequality,
using the fact that $w$ vanishes on the boundary of $D(0)$. 
Thus $\| |v| |\nabla w | \|_1, \|w\|_1\le \|\nabla w\|_1$.
The claim of the proposition follows by applying  
Lemma 3.5 
to $w$. 
\end{proof}

\begin{proof}[Proof of Theorem 3.8]
Let $c_1 = 4$, $c_2 = 2$. 
The l.h.s.
 is bounded by $\| x^{(L)}  - D_\tau x^{(L)} \| + \| D_\tau x^{(L)}[x^{(0)}] - x^{(L)}[D_\tau x^{(0)}] \| $.
The second term is less than $2 c_1 L | \nabla \tau |_\infty \|x^{(0)}\|$ by Theorem  3.7. 
To bound the first term, 
we apply Proposition 3.4, 
and notice that for all $\lambda', \lambda$, 
$\| \nabla W^{(L)}_{\lambda' , \lambda}\|_1 \le 2^{-j_L} \pi \| a^{(L)}_{\lambda, \lambda}\|_{FB}$ 
(consider 
$W^{(L)}_{\lambda' , \lambda}(u) = W(u) 
= \sum_k a_k \psi_{J, k}(u) = 2^{-2 J} \sum_k a_k \psi_{k}(2^{- J}u)$,  $J=j_L$,
let $w(u) = \sum_k a_k \psi_{k}(u)$, 
then $W(u) = 2^{-2 J} w(2^{-J}u)$,
and $ \| \nabla W\|_1 = 2^{-J} \| \nabla w\|_1$, where $\| \nabla w\|_1 \le \sqrt{\pi} \|a\|_{FB}$ by Lemma 3.5), 
and thus $D_L \le 2^{-j_L}  A_L$. 
By (A2'),
this gives that $\| D_\tau x^{(L)} - x^{(L)}  \|  \le  c_2 2^{-j_L}  |\tau|_\infty  \|x_c^{(L-1)}\|$,
and 
$ \|x_c^{(L-1)}\| \le \|x^{(0)}\|$ by Proposition 3.1 
(b).
\end{proof}

\section{Experimental Details}

\begin{figure}[h]
\begin{center}
\includegraphics[width = 0.45 \linewidth]{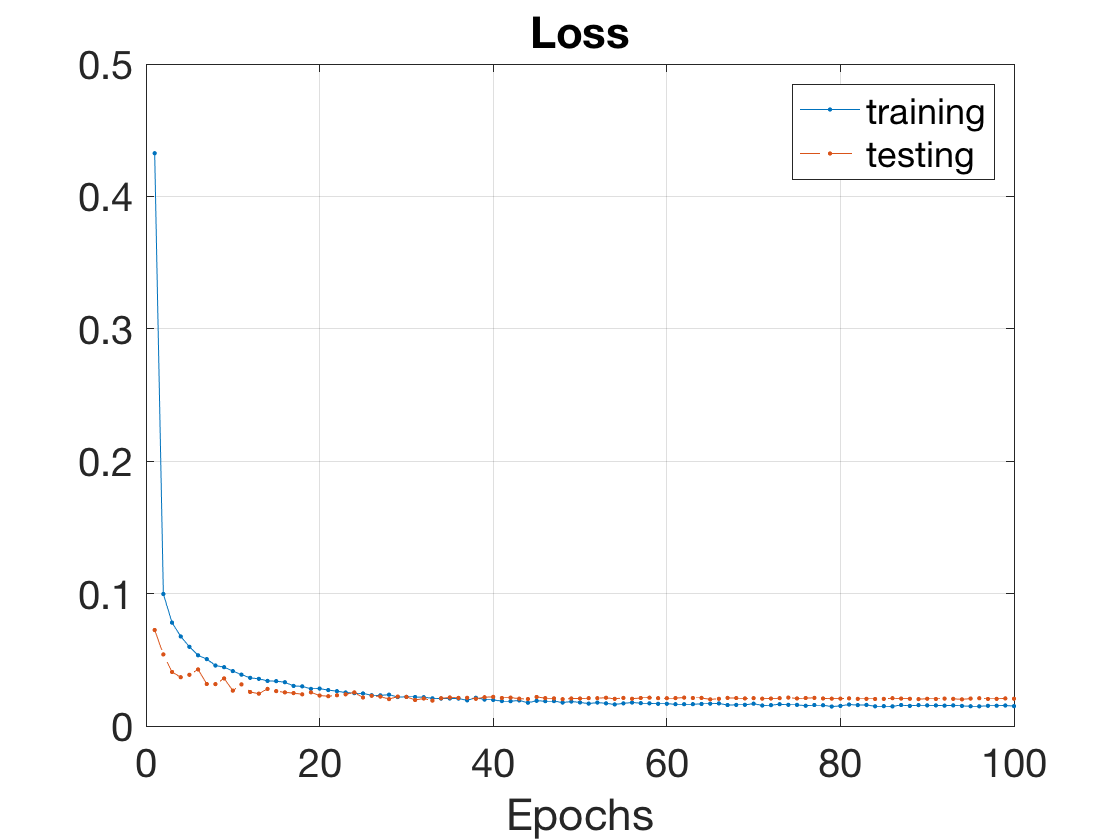} 
\includegraphics[width = 0.45 \linewidth]{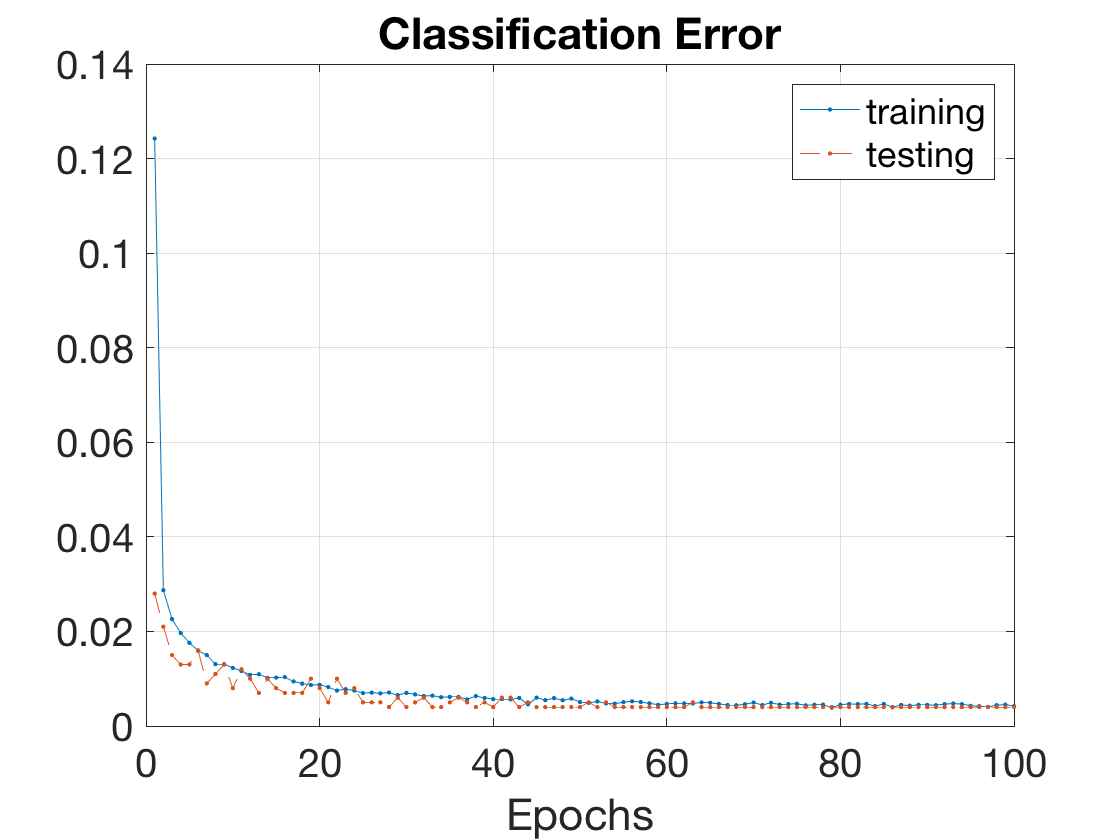} 
\caption{
The evolution of training and validation losses (left) and errors (right)
 over the epochs 
of a 
Conv-2 DCF-FB network trained on 50K MNIST using SGD.
}
\label{fig:train}
\end{center}
\end{figure}

The training of a Conv-2 DCF-FB network (Table 2) on MNIST dataset: 

The network is trained using standard Stochastic Gradient Descent (SGD) with momentum $0.9$ and batch size $100$ for 100 epochs. 
$L^2$ regularization (``weightdecay") of $10^{-4}$ is used on the trainable parameters $a$'s.
The learning rate decreases from $10^{-2}$ to $10^{-4}$ over the 100 epochs.
Batch normalization is used after each convolutional layer.
The typical evolution of training and testing losses and errors over epochs are shown in Figure \ref{fig:train}.

\end{document}